# An Artificial Immune System Model for Multi Agents Resource Sharing in Distributed Environments

Tejbanta Singh Chingtham
Dept. of Computer Sc. & Engineering
Sikkim Manipal Institute of Technology,
Majitar, Rangpo, Sikkim 737132 India

G. Sahoo
Dept. of Information Technology
Birla Institute of Technology
Mesra, Ranchi, Jharkand
India

M.K. Ghose
Dept. of Computer Sc. & Engineering
Sikkim Manipal Institute of Technology,
Majitar, Rangpo, Sikkim 737132 India

**ABSTRACT**
Natural Immune system plays a vital role in the survival of the all living being. It provides a mechanism to defend itself from external predates making it consistent systems, capable of adapting itself for survival incase of changes. The human immune system has motivated scientists and engineers for finding powerful information processing algorithms that has solved complex engineering tasks. This paper explores one of the various possibilities for solving problem in a Multiagent scenario wherein multiple robots are deployed to achieve a goal collectively. The final goal is dependent on the performance of individual robot and its survival without having to lose its energy beyond a predetermined threshold value by deploying an evolutionary computational technique otherwise called the artificial immune system that imitates the biological immune system.

## 1. INTRODUCTION

In recent years there has been considerable interest in exploring and exploiting the potential of biological systems for applications in computer science and engineering. These systems are inspired by various aspects of the immune systems of mammals. Artificial immune system imitates the natural immune system that has sophisticated methodologies and capabilities to build computational algorithms that solves engineering problems efficiently [2]. The main goal of the human immune system is to protect the internal components of the human body by fighting against the foreign elements such as the fungi, virus and bacteria [1]. Moreover, research into natural immune systems suggests the existence of learning properties which may be used to advantage in machine learning systems [5].

Similarly, if there is an environment which is divided into sub environment then each sub environment is traversed by a single bot. Every bot is assigned to do a set job in its environment. Considering an environment being divided into n sub environment with m Bots, each working on one environment, the complete environment may be obtained by summing up all the individual bot and the sub-environment The objective of this research is to demonstrate the utility of multi-robot deployed using a unique First Come First Serve (FCFS) charging where only a single charger is used by multiple bots in an environment such that none of the bots are allowed to stop functioning by complete discharge of the battery power. To achieve this unique goal a new computational technique called the Artificial Immune System is applied which presumes the discharge of power of the battery as an external attack to malign the operation of the robot in the environment and uses natural immune concepts to make the robot immune to such failure.

## 2. IMMUNE SYSTEM

The immune system defends the body against harmful diseases and infections. It is capable of recognizing virtually any foreign cell or molecule and eliminating it from the body. To do this, it must perform pattern recognition tasks to distinguish molecules and cells of the body called "self" from foreign ones called "non self". Thus, the problem that the immune system faces is that of distinguishing self from dangerous non self [1]. Antibodies which are also referred to as immunoglobulin are Y-shaped proteins that respond to a specific type of antigen like bacteria, virus or toxin that contain a special section at the tip of the two branches of the Y that is sensitive to a specific antigen and binds to it. When an antibody binds to a toxin it becomes an antitoxin and normally disables the chemical action of the toxin [6]. Based on a study of the human immune system, we have drawn some properties that can serve as design principles of artificial immune based multi agent systems. The properties relevant to the proposed model are discussed below.

Immune memory: It is a result of clonal expansion. Some of the cloned cells differentiate into memory cells and the rest of the clones become plasma cells.

Jerne's idiotopic network deals with the interaction of antibodies. Jerne's network is a network of B cells that communicate the shape of the antigenic epitope amongst them through idiotopes and paratopes [2].

A huge amount of antibodies can bind to an invader and then it signals the complement system [7] that the invader needs to be removed. Antigens are defective coding on the cell surface that appears soon after the infection of a cell by an infectious agent, but before replication has begun. Epitopes, which are patterns, present on the surface of the antigen are used by the antibody to detect if they constitute a potential threat to the body. When the Paratope of an antibody matches the Epitope of the antigen, a reaction to suppress the antigen is initiated. In case the match is not exact, the antibody undergoes a process called somatic hypermutation [6], a controlled version of mutation, to set it right. The immune system is unique, robust, autonomous and multi-layered. It is augmented with a distributed learning mechanism having lasting memory [7]. This shows the overall functioning of the immune system. The immune system recognizes the antigens and the antigenic patterns are identified. On identification of an antigenic pattern, the B cells communicate the information in parallel to each other by means of paratopes and idiotopes in the network [3].

## 3. ARTIFICIAL IMMUNE SYSTEM

The immune system is highly complicated and appears to be precisely tuned to the problem of detecting and eliminating infections. It is believed that it also provides a compelling example of a distributed information-processing system, one which we can study for the purpose of designing better artificial adaptive systems [3], [4]. AIS uses the concepts of natural Immune System to improve the computational techniques and in this paper an attempt is made to provide mechanisms to prevent failure in a multi-robot environment. Every autonomous robot works on limited power and it is drained as the robot works on its course. Using the AIS concept the loss of energy or power from the robot is viewed as an antigen and immune concepts are incorporated to ensure that such antigen do not disrupt the normal functioning of any of the robots.

When more then one bot work autonomously in an environment, it is made sure that none of the bot is allowed to lose its power completely making it non functional. If any one of the bot has weak battery strength that is below the threshold value and is waiting in a long queue to recharge itself, then it may stop functioning completely by it is allowed to be recharged. In order to avoid such eventualities and making one of the robot invalid this simulation allows such robots with critical threshold values to jump the queue to charge immediately ignoring the normal rules of the queue. If such critical cases are not encountered the robots follows the normal characteristics of a queue, thereby ensuring that no robots are allowed to be invalid and completes the assigned task. This suggested technique makes a multi-robot or multi-agent scenario more robust and consistent ensuring the completion of a desired goal.

## 4. ENVIRONMENT DESCRIPTION

Every bot traverses its environment to do a set job. Considering an Environment(E) being divided into n sub environments $g(E_n)$ where n is the no of the environment (n = 1, 2, 3, 4....n). The work done by each of the bot is $w(E_n)$. So by summing up the work done by a bot on a sub environment gives the work done on an environment.

$W(E) = w(E_1)g(E_1) + w(E_2)g(E_2) + w(E_3)g(E_3) + w(E_4)g(E_4) + \ldots$

$$+ w(E_n)g(E_n)$$

### 4.1 Working Description

The algorithm followed is normal FCFS. When there are lot of variables waiting in a queue then an algorithm is followed in which the first variable to have entered the waiting queue will be the first one to come out of the queue. Whenever the battery is running low then the bot stores its present position and moves towards its docking point. On reaching its docking point the bot check s for the charger. If the charger is empty it moves to the charger otherwise it enters in the queue while waiting for its turn to avail the charger .in this case it follows FCFS. In case the bot is running very low and is towards the rear of the queue then it is made to jump to the top of the queue, hence being the first to avail the charger in the queue. This way the bots are saved from complete failure due to the lack of battery charge. After the charging is complete the bot moves back the position where it had encountered its low power problem. It then resumes its previous work that it was doing.

*4.1.1 Simulation Workspace*

There are four Autonomous Bot working in different sub environment. Each bot starts from any random position & battery strength. Once the bot reaches the red mark it moves back to the green mark in a straight line. There is only one charger shared by all the four bot. Each bot has a unique docking (waiting) station in case the charger is in use.

The following are the battery strength indicator

| High | Average | Medium | Low | Very low |
|---|---|---|---|---|
| 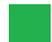 | 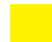 | 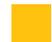 | 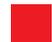 | 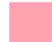 |
| Green | Yellow | Orange | Red | Pink |

The environment considering (n=4) where n is the no of bots and which is also the no of sub environments is described from Fig 1 through Fig 11.

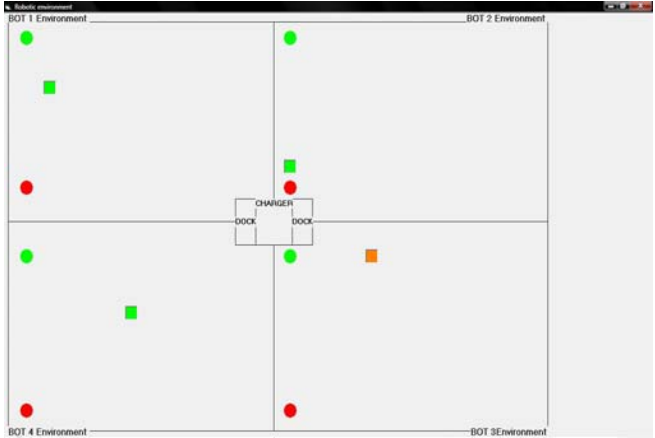

Fig 1. All the bot start from any random position with any random battery strength.

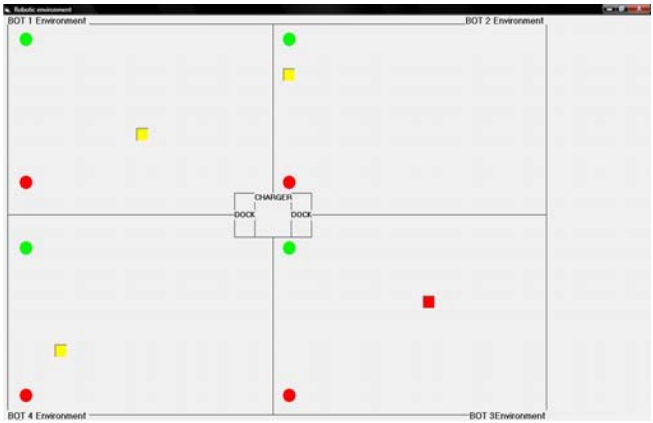

Fig 2. All the bot is working while Bot3 is on low battery.

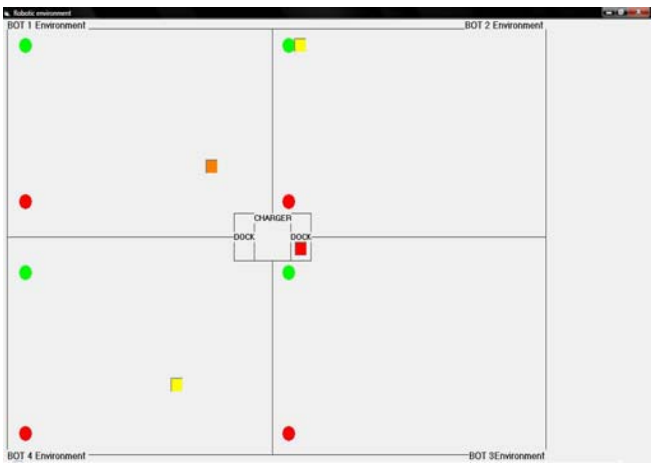

Fig 3. Bot 3 reaches its docking station and checks for the charger's availability.

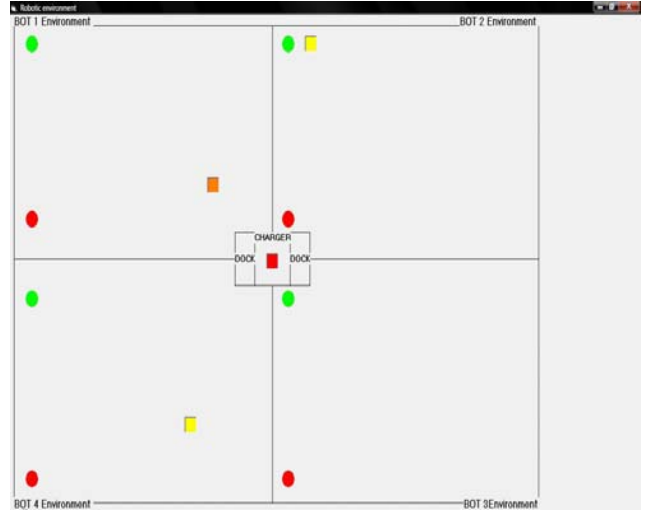

Fig 4. Bot 4 moves to the charger as the charger is not in use and no other bot is there in the docking station.

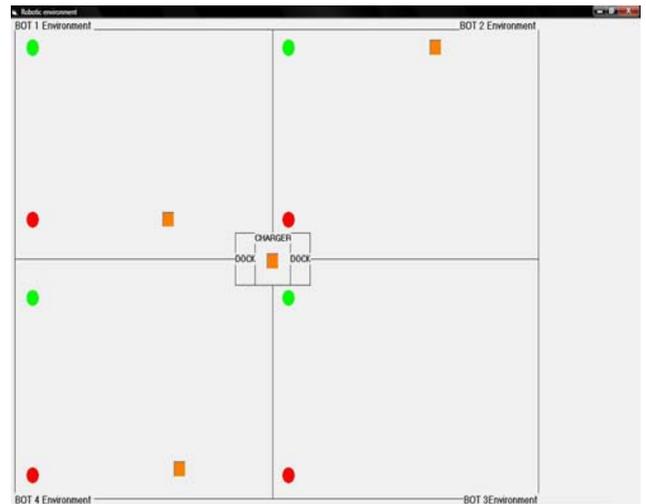

Fig 5. Bot 4 starts charging its battery.

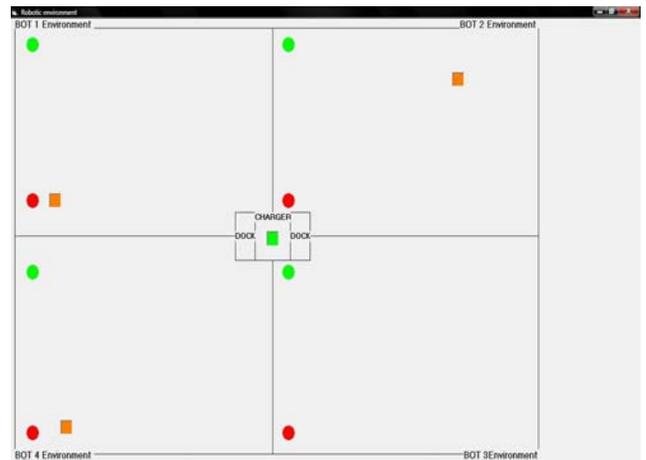

Fig 6. The bot eventually gets charged starts moving to its position.

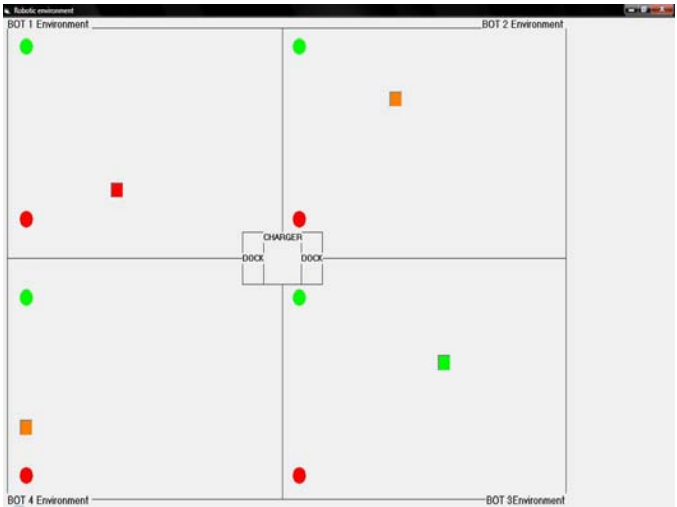
Fig 7. The bot comes back to the the position where it had last left from for charging.

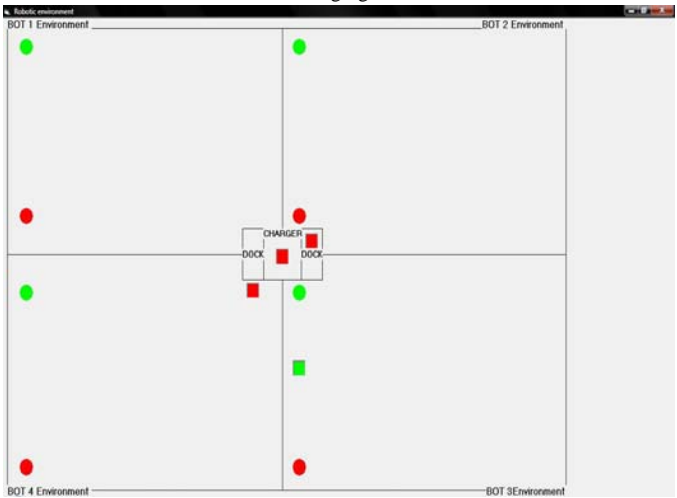
Fig 8. Bot 1 is being charged, Bot 2 is waiting next in the queue for the charger, Bot 4 is just going to reach the dock for charging.

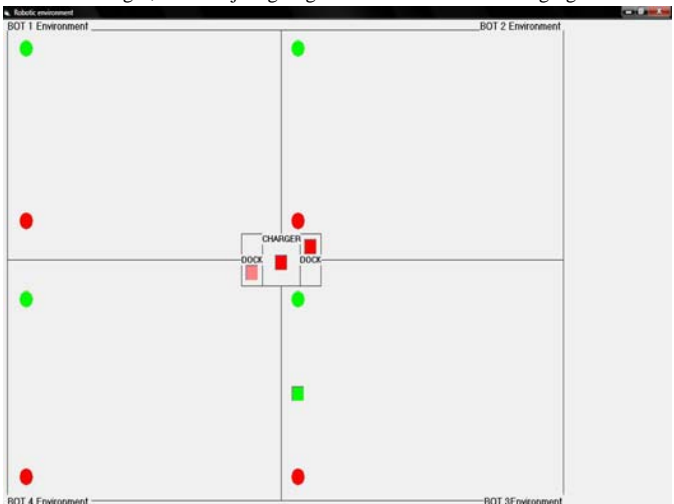
Fig 9. Bot 4 has reached the lowest battery phase

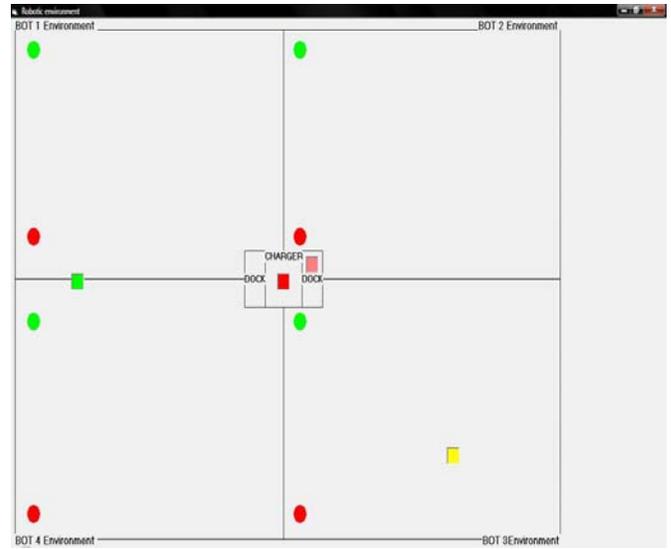
Fig 10. Bot 1 has finished charging, Bot 4 goes for the charger while the Bot 2 waits reaching its lowest battery phase.

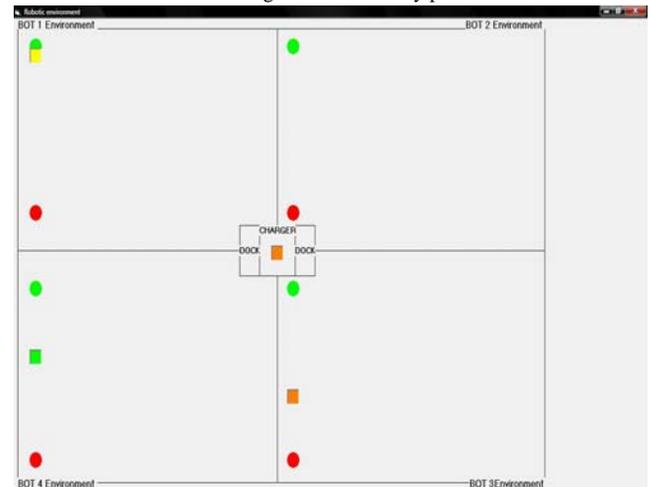
Fig 11. After Bot 4 is finished charging Bot 2 goes for the charger

In Fig 1 It is seen that the bots starts traversing from any random initial point with any random battery strength. The bots keep exploring their environment in a predefined way. Once any one of the bot encounters a weak battery problem it stores its present position and moves to its docking point as shown in Fig 2 and Fig 3. On reaching its docking point it checks for the availability of the charger, if the charger is available then it moves to the charger and starts charging as shown in Fig 4 and Fig 5. Once the bot is charged it moves to its last stored working position as elaborated in Fig 5 and Fig 6.

Considering more then two bots in their docking station and one in the charger, the first bot is considered as the next bot for charging as in FCFS described in Fig 7 and hence according to the Fig 7, bot 2 should be going in next to the charger. However in Fig 8, it is seen that bot 4 battery has become very weak so it is given a jump in the queue before bot 2 which was considered to go in next. So bot 4 gets the charger before bot 2 as in Fig 9. At last after bot 4 is charged bot 2 is charged as shown in Fig 10.

*4.1.2 BotCharging Algorithm*

*Start*
 *Loop: a = 1*
   *For (i = 1 to 4)*
     *Start bot(i) from any random position with any random battery strength*
     *bot(i).position = bot(i).position + 1 Centimeter*
     *bot(i).battery strength =*
     *bot(i).battery strength – 1 Micro charge unit*
     *if(bot(i).battery strength < $2^{nd}$ last battery strength)*
       *Call Charging*
     *End if*
     *If bot(i).position = red post*
       *Move to Green post*
     *End if*
   *Exit for*
   *Goto Loop*
 *End*
*Start sub Charging*
   *X(i)= bot(i).position*
   *bot(i).position = bot(i).position + 1*
   *if (bot(i).position <> bot(i).dock)*
     *Call Priority FCFS*
   *End if*
*End sub*
*Start sub Priority FCFS*
   *If (charging.position = empty)*
     *bot(i).position = charging.position*
     *charging.position = nonempty*
   *Else*
     *Queue(a)=i*
     *a = a+1*
     *If bot(i).strength = last battery strength*
       *const = Queue(a)*
       *i = a*
       *For (c =i to 1)*
         *Queue(c)= Queue(c-1)*
         *c = c – 1*
       *Exit for*
     *End if*
   *End if*
*End sub*

## 5. CONCLUSION

This research attempts to model a simulation environment based on artificial immune system applicable to intelligent multi agents [3]. An application for the model is simulated. None of the bot is allowed to die by implementing artificial immune system. A unique FCFS is implemented which is re-scheduled if the battery strength of one of the bot goes below the threshold level. The bot whose strength goes down this level is made to jump to the top of the queue, thereby increasing the consistency and efficiency of the complete system.

## 6. ACKNOWLEDGMENTS

This work is in part supported by AICTE-RPS grant vide Grant No 8023/BOR/RID/RPS-235/2008-09

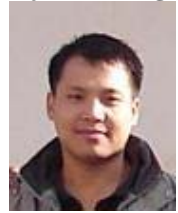
**Tejbanta Singh Chingtham** is associated with Sikkim Manipal University since August, 2000 and is working as an Associate Professor in the Department of Computer Science & Engineering. He received his B.Tech in Computer Science & Engineering from Bharathiar university, Coimbatore, India and M.Tech from Indian Institute of Technology Guwahati, India in Computer Science & Engineering and pursuing PhD from Birla Institute of Technology Mesra, Ranchi, India. He is a member of IEEE, IAENG and IACSIT. He has served as Technical Committee Members of Various Journals and Conferences. His main areas of Interest are Artificial Immune System, Natural Computing, Evolutionary Computation and Mobile Autonomous Robots, Navigation and Path Planning.

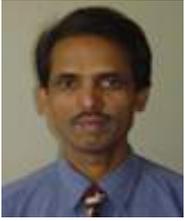

**G. Sahoo** received his M.Sc. degree in Mathematics from Utkal University in the year 1980 and Ph.D. degree in the area of Computational Mathematics from Indian Institute of Technology, Kharagpur in the year 1987. He has been associated with Birla Institute of Technology, Mesra, Ranchi,India since 1988 and is currently working as Professor and Head in the Department of Information Technology. His research interest includes Theoretical Computer Science, Parallel and Distributed Computing, Evolutionary Computing, Information Security, Image Processing and Pattern Recognition.

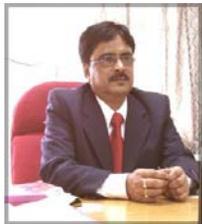

**M.K.Ghose** obtained his Ph.D. from Dibrugarh University, Assam, India in 1981. He is currently working as the Professor and Head of the Department of Computer Science & Engineering at Sikkim Manipal Institute of Technology, Mazitar, Sikkim, India. Dr. Ghose also served in the Indian Space Research Organization (ISRO) – during 1981 to 1994 at Vikram Sarabhai Space Centre, ISRO, Trivandrum in the areas of Mission simulation and Quality & Reliability Analysis of ISRO Launch vehicles and Satellite systems and during 1995 to 2006 at Regional Remote Sensing Service Centre, ISRO, IIT Campus, Kharagpur(WB), India in the areas of RS & GIS techniques for the natural resources management. Dr. Ghose has conducted a number of Seminars, Workshop and Training programmes and published around 95 technical papers in various national and international journals in addition to presentation/ publication of 125 research papers in international/ national conferences. He has supervised PhD desertations in areas like data mining, information security, geoinformatics and Robotics. He is a Life Member of Indian Association for Productivity, Quality & Reliability, Kolkata, National Institute of Quality & Reliability, Trivandrum, Society for R & D Managers of India, Trivandrum and Indian Remote Sensing Society, IIRS, Dehradun